\documentclass[10pt,twocolumn,letterpaper]{article}
\pdfoutput=1
\usepackage{wacv}
\usepackage{times}
\usepackage{epsfig}
\usepackage{framed}
\usepackage{graphicx}
\usepackage{amsmath}
\usepackage{amssymb}
\usepackage[table]{xcolor}
\usepackage{multirow}
\usepackage{hyperref}
\usepackage{listings}
\usepackage{cuted}
\usepackage{capt-of}
\definecolor{shadecolor}{rgb}{0.95,0.95,0.92}

\usepackage{algorithm,algpseudocode}

\usepackage{pifont}
\definecolor{shadecolor}{rgb}{0.95,0.95,0.92}
\definecolor{codegreen}{rgb}{0,0.6,0}
\definecolor{codegray}{rgb}{0.5,0.5,0.5}
\definecolor{codepurple}{rgb}{0.58,0,0.82}
\definecolor{backcolour}{rgb}{0.95,0.95,0.92}

\newcommand{\Angie}[1]{{\color{black} #1}}
\lstdefinestyle{mystyle}{
    commentstyle=\color{codegreen},
    keywordstyle=\color{magenta},
    numberstyle=\tiny\color{codegray},
    stringstyle=\color{codepurple},
    basicstyle=\ttfamily\footnotesize,
    breakatwhitespace=false,
    breaklines=true,
    captionpos=b,
    keepspaces=true,
    numbers=left,
    numbersep=5pt,
    showspaces=false,
    showstringspaces=false,
    showtabs=false,
    tabsize=2
}

\newtheorem{mydef}{Definition}

\wacvfinalcopy 

\def\wacvPaperID{1063} 

\ifwacvfinal\fi
\begin{document}

\title{Dynamic Spectral Residual Superpixels}

\author{Jianchao Zhang$^{1*}$ , Angelica I. Aviles-Rivero$^{2*}$, Daniel Heydecker$^{2}$\thanks{Equal Contribution. }, \\ Xiaosheng Zhuang$^1$, Raymond Chan$^1$ and Carola-Bibiane Sch{\"o}nlieb$^2$
   \\
$^1$ City University of Hong Kong (CityU), Hong Kong.  \\
$^2$ DPMMS and DAMTP, University of Cambridge, UK. \\
{\tt\small jzhang537-c@my.cityu.edu.hk, \{ai323,dh489,cbs31\}@cam.ac.uk, \{xzhuang7,rchan.sci\}@cityu.edu.hk  }
}

\maketitle
\ifwacvfinal\thispagestyle{empty}\fi

\begin{strip}\centering
\includegraphics[width=\textwidth]{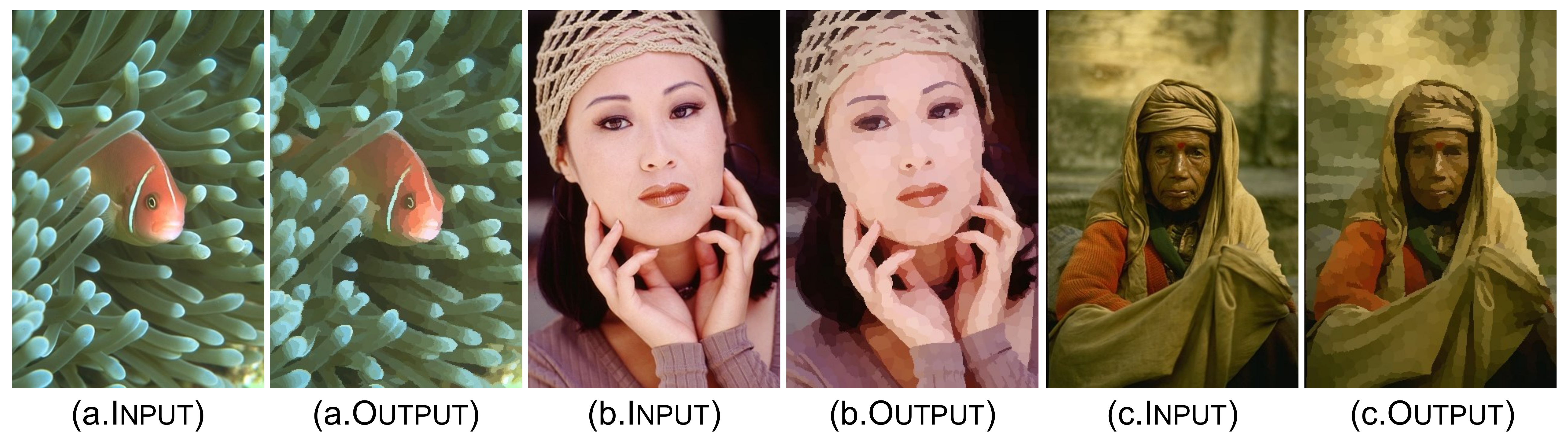}
\captionof{figure}{Input-Output examples of our proposed approach for three images with different characteristics such as diverse  structure and colours. Our approach seeks to adhere better to the boundaries through a structure measurement whilst connecting meaningful regions, for example see the eyes and mouth at the three outputs.}
\label{fig:teaser}
\end{strip}

\begin{abstract}
We consider the problem of segmenting an image into superpixels in the context of $k$-means clustering, in which we wish to decompose an image into local, homogeneous regions corresponding to the underlying objects. Our novel approach builds upon the widely used
Simple Linear Iterative Clustering (SLIC), and incorporate a measure of objects' structure based on the spectral residual of an image. Based on this combination, we propose a modified initialisation scheme and search metric, which helps keeps fine-details. This combination leads to better adherence to object boundaries, while preventing unnecessary segmentation of large, uniform areas, while remaining computationally tractable in comparison to other methods. We demonstrate through numerical and visual experiments that our approach outperforms  the state-of-the-art techniques.
\end{abstract}

\section{Introduction}
Image segmentation has been a widely explored task in computer vision yet a still open problem. In particular, superpixels segmentation has become a pre-processing tool for several applications including classification~\cite{sellars2019superpixel,shi2019multiscale},  optical flow~\cite{menze2015object,liu2019selflow}, colour transfer~\cite{giraud2017superpixel}, depth estimation~\cite{liu2015deep,chen2018accurate} and tracking~\cite{wang2011superpixel,yeo2017superpixel}. The central idea of superpixels is to split a given image in multiple clusters, which reflect semantically meaningful regions.

There are several advantages of using superpixel representation instead of working at pixel-wise level. Firstly, an application becomes computationally and representationally efficient as the number of primitives are significantly reduced. Secondly, the natural redundancy in an image is exploited, and therefore, features can be extracted on representative regions whilst reducing noise and increasing discriminative information~\cite{ren2003learning,achanta2012slic,stutz2018superpixels}.

Since the pioneering work of Ren and Malik~\cite{ren2003learning}, the community has devoted to develop different algorithmic approaches to improve over~\cite{ren2003learning}.  These approaches can be roughly divided in: graph-based methods e.g.~\cite{ren2003learning, humayun2015middle}, path-based approaches e.g.~\cite{tang2012topology}, density-based models e.g.~\cite{vedaldi2008quick}, contour models e.g.~\cite{levinshtein2009turbopixels} and clustering methods e.g.~\cite{achanta2012slic,maierhofer2018peekaboo}.

Out of all of the approaches reported in the literature, the Simple Linear Iterative Clustering (SLIC)~\cite{achanta2012slic} is perhaps the most popular method that offers a good performance whilst demanding low computational cost, by building on Lloyd's algorithm~\cite{lloyd1982least} for $k$-means. The central idea of SLIC is to perform the superpixels partition based on an iterative scheme that search for similarities between points ensuring at each step that we assign points to the nearest cluster from the previous step.

The ability of SLIC to obtain a good segmentation with low computational cost comes from the observation that, by using a similarity metric, one can greatly reduce the number of distance calculations required. However, SLIC is also limited by its own construction, and in particular, by its search range, and one can thus observe two major limitations. Firstly, SLIC tends to segment large uniform regions in an image with more superpixels than are intuitively necessary, which limits resolution in other parts of the image. Secondly, in structure-rich parts of the image, the final superpixel size is much smaller than the search radius of SLIC, which leads to many unnecessary distance computation. Finally, since we expect structure-rich parts of the image to have a higher density of superpixels, it may be efficient to perform the initial seeding of cluster centres in anticipation of this inhomogeneity.

In this work, we propose a new algorithmic approach, exhibited in Fig.~\ref{fig:teaser}, that improves upon the SLIC approach, motivated by the drawbacks discussed above. We show that our approach outperforms SLIC and several works on the body of literature. Our main contributions are as follows.

\begin{itemize}
\itemsep0em
    \item We propose a new superpixel approach, which incorporates the \emph{saliency function} $\mathcal{S}(x)$ of Hou et al. \cite{hou2007saliency} as a proxy for object density. This leads to the following advantages.
        \begin{itemize}
        \itemsep0em
            \item By incorporating the saliency $\mathcal{S}(x)$ into the distance computation, we can prevent unnecessary over-segmentation of large, uniform regions, such as the sky in the first example of Fig. \ref{fig::compStructuresv5}, and allowing greater focus on structure-rich parts of the image.
            \item We propose a new seeding strategy, based on the inhomogenity described by $\mathcal{S}$. This allows for greater resolution changes at fewer iterations by focusing on relevant structures, and hence keeping fine-details of the structures in the final segmentation.
        \end{itemize}
    \item We extensively evaluate our approach with a large range of numerical and visual experiments.
    \item We demonstrate that our two major contributions mitigates the major drawbacks of the state-of-the-art techniques, by reporting the lowest undersegmentation error and highest boundary recall.
\end{itemize}

\color{black}
\section{Related Work}
In this section, we review the body of literature in turn. We then highlight the advantages of clustering based methods, and their current drawbacks that motivate our new algorithmic approach. \medskip

There have been different attempts in the literature to improve superpixels segmentation. A set of approaches tackle the problem using graph representation of the image and the partition is based on the similarity of the nodes, e.g. colour, including~\cite{ren2003learning, shen2014lazy,shi2000normalized,felzenszwalb2004efficient,moore2008superpixel,yang2013saliency}. However, although promising results are reported, the computational time is often very high. Another perspective has been followed by local mode-seeking algorithms including the well-known Quick Shift, which partition is based on an approximation of kernelised mean-shift~\cite{vedaldi2008quick}. However, there is not control on the number of superpixels or compactness.

Another set of approaches addressed the superpixel partition problem as the task of finding the  shortest path between seeds, for instance using the the well-known Dijkstra algorithm, as reported in~\cite{tang2012topology,fu2014regularity}. However, this type of approach is usually unable to control the compactness. We briefly mention other methods for superpixels segmentation. A body of work has proposed algorithms for image segmentation based on \emph{gradient ascent} and other geometric methods \cite{comaniciu2002mean,vincent1991watersheds,levinshtein2009turbopixels}. For an extensive review of the literature, we refer the reader to ~\cite{stutz2018superpixels}.

In particular, in this work we focus on, probably, the most popular  superpixel category, which is clustering based approaches. The basis of this perspective builds on Lloyd's algorithm~\cite{lloyd1982least} for $k$-means clustering. The main idea of this algorithm is to  partition a set of observations into $k$ clusters, in which each observation is assigned to the cluster with the nearest mean, and produces excellent results at the cost of high computational intensity.
Within this category, one can find a top reference approach called Simple Linear Iterative Clustering (SLIC).  SLIC was proposed by Achanta et al. \cite{achanta2012slic}, in which authors propose a local version of the Lloyd's algorithm, which is computationally much simpler while keeping excellent segmentation quality.

Following this philosophy, different algorithmic approaches have been proposed including~\cite{wang2012vcells,papon2013voxel,neubert2014compact,li2015superpixel}. Most recently, in~\cite{achanta2017superpixels} authors proposed an improved version of SLIC, in which they proposed to compute a polygonal partition to adapt better to the geometry of the objects in the image.  Maierhofer et al~\cite{maierhofer2018peekaboo} propose a \textit{dynamic refinement} of this method, called dSLIC, which seeds the initial cluster points inhomogenously and allows the search radius to vary across clusters, both according to a measure of local object density. This allows better capturing of fine details in structure-rich regions, and further reduces computational complexity by eliminating unnecessary searches.

Let us also mention the closely related problem of \emph{salient object detection}. In this problem, one has the simpler goal of identifying which regions in an image contain \emph{salient} or novel information, and which contain only patterns and structures repeated throughout the image. This problem shares some similarities with the problem of image segmentation; for instance, one might hope that the salient objects are identified as superpixel regions. A hugely successful method in this problem, based on Fourier analysis, was proposed by Hou et al. \cite{hou2007saliency}, which inspires our current approach. More recent works include techniques based on graphs \cite{yang2013saliency} or machine-learning \cite{li2016deep,tong2015salient}.

\section{Proposed Approach}
In this section, we describe in detail our superpixel approach. Firstly, we formalise the definition of superpixels in terms of a clustering task. We then introduce the details of both our new measure of structure function and our initialisation strategy.
\medskip

We view an input image, of width A and height B, as a map $I:\mathcal{X}\rightarrow \Omega$, where $\mathcal{X}=\mathcal{[A]}\times\mathcal{[B]}$ is a rectangular domain, and $\Omega\subset \mathbb{R}^3$ appropriate colour domain. We also define a metric $d$ on $\mathcal{X}\times \Omega$, representing the similarity of points in space and with different colour values, and a \emph{feature map}
$\mathcal{F}$, which takes a subset $S\subset \mathcal{X}$ and returns a pair in $\mathcal{X}\times\Omega.$ \emph{$k$-means clustering} now seeks a partition of $\mathcal{X}$ into path-connected sets $\{S_i\}_{i=1}^n$ such that, for each $i$, $S_i$ is exactly the set of points where the infimum $\inf_j d((x,I(x)),\mathcal{F}(S_j))$ is attained at $j=i.$

\begin{figure*}[t!]
\centering
\includegraphics[width=1\textwidth]{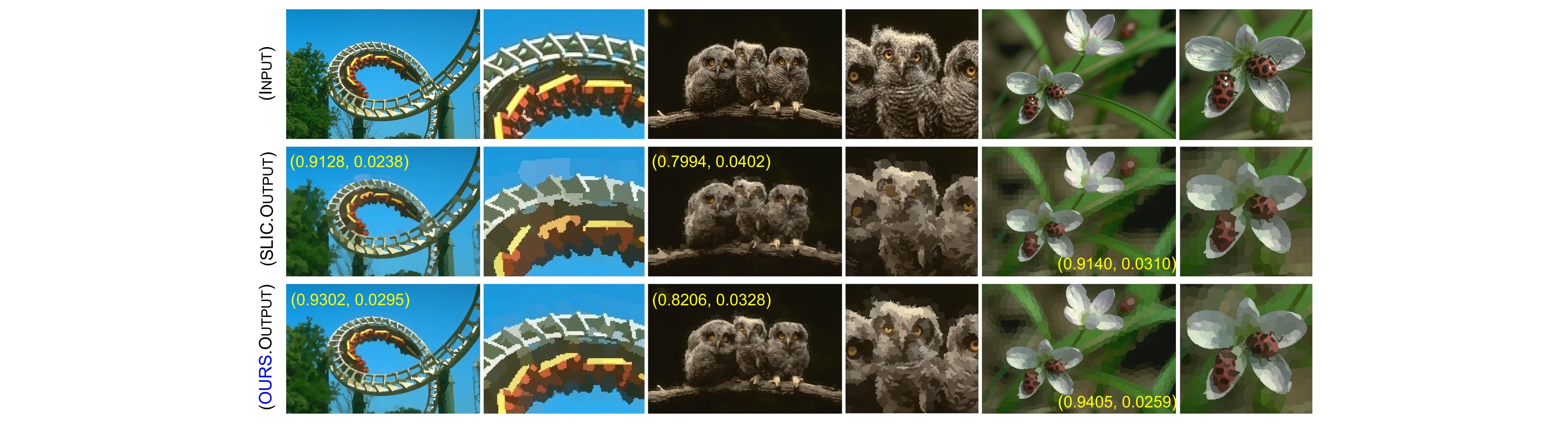}
\caption{Illustration output of our approach against SLIC. SLIC tends to over-segment uniform areas with more superpixels than necessary, such as the sky in the first image, and fails to preserve fine-structures, such as the owl's eyes and the  roller coaster at the zoom-in views. The numerical results in parenthesis denote (Undersegmentation Error, Boundary Recall). The lower the undersegmentation error the better whilst the higher the boundary recall the better.}
\label{fig::compStructuresv5}
\end{figure*}

\subsection{Object Density Measure via Spectral Residual}
The key strength of dSLIC \cite{maierhofer2018peekaboo} over SLIC is the recognition that objects in an image are not distributed uniformly, and that image segmentation can exploit this to improve segmentation results and computational efficiency. Our approach is to exploit this same principle further, and use the strength of the \emph{Spectral Residual} approach proposed by \cite{hou2007saliency} as a better measure of object detection.
\smallskip

We briefly review the Fourier analysis leading to the definition of the spectral residual in \cite{hou2007saliency}. For an image $I$, we write $\mathfrak{F}I$ for the Fourier transform, which is a matrix of the same dimensions as $I$, and whose arguments we will write as two-dimensional frequencies $f$. The log-spectrum of an image $I$ is then given by \begin{equation}
\mathcal{L}(f)=\log(\mathfrak{R}(\mathfrak{F}I)(f))
\end{equation}

\noindent
where $\mathfrak{R}$ denotes the real part; we also write $\mathcal{P}(f)=\mathfrak{I}(\mathfrak{F}I)(f)$ for the imaginary part, or phase spectrum. The key insight of \cite{hou2007saliency} is that much of the information contained within $\mathcal{L}$ is redundant, because $\mathcal{L}$ is, to a good approximation, locally linear. These features are then encapsulated in the local average $\mathcal{A}(f)=(h_n\star \mathcal{L})(f)$, where $h_n$ is the matrix consisting entirely of $\frac{1}{|\mathcal{X}|}$, and the \emph{residual} log-spectrum, corresponding to the salient features, is given by
\begin{equation}
\mathcal{R}(f)=\mathcal{L}(f)-\mathcal{A}(f)=\mathcal{L}(f)-(h_n\star \mathcal{L})(f).
\end{equation}

The final saliency map, which we take as our measure of object density, is then given by recombining with the phase spectrum, inverting the Fourier transform and adjusting the resulting function. Therefore, our proposed function reads:
\begin{equation}
    \mathcal{SR}(x)=g_\sigma \star \left(\mathfrak{F}^{-1}\left\{e^{\mathcal{R}(f)+\mathcal{P}(f)}\right\}\right)^2(x),
    \label{eq:structureFunction}
\end{equation}

\noindent
where the squaring ensures that the quantity considered is nonnegative, and the convolution with a Gaussian kernel $g_\sigma$ ensures that the final result is smooth. For our purposes, we found that in practice $\sigma=20$ is an excellent value. We then set a rescaling step for the search radius using~\eqref{eq:structureFunction} as $\mathcal{G}(x):= \exp(\mathcal{SR}(x)-\overline{\mathcal{SR}})$, where $\overline{\mathcal{SR}}$ denotes the mean of the structure function on the image grid. We then propose to have the distance computations depending on our function, which reads:
\begin{shaded*}
\vspace{-0.5cm}
\begin{lstlisting}[language=Python, mathescape=true]
#pseudocode for  computing distances
input $m$ (compactness), number of superpixels
while $e_{r}$ $\leq$  outset
    for  $x\in\mathcal{X},1\leq i\leq k$:
        if $|x-(\mathcal{F}(S^{(t)}_i))_1|\leq 2S\mathcal{G}(\mathcal{F}(S^{(t)}_i))$:
            #distance computation
            Compute: $d((x,I(x)),\mathcal{F}(S^{(t)}_i))$
        else:
            $d((x,I(x)),\mathcal{F}(S^{(t)}_i))=\infty$
$S_i^{t+1}= \min_{1\leq j \leq n}d((x,I(x)),\mathcal{F}(S^{(t)}_j))$
Compute Residual Error $e_{r}$
Increase t=t+1
\end{lstlisting}
\vspace{-0.5cm}
\end{shaded*}

By incorporating our proposed function, which we use as a measure of object density, into the distance computation one obtains have two major advantages. Firstly, by doing a dynamic adjustment of the search range based on our function $\mathcal{G}$, one can connect uniform regions, and so avoid segmenting the images into unnecessary small superpixels. This effect is illustrated in Fig.~\ref{fig::compStructuresv5}, for example, at the second column where the our approach was able to keep the sky in a same region, and the yellow car. Secondly, our approach focuses on segmenting fine details by capturing relevant structures; this is visible in the owl's eyes and head in Fig.~\ref{fig::compStructuresv5}.Furthermore, we remark that our approach does not introduce additional parameters to fine-tuning.

We now turn to explain our second major modification, which concerns the seeding initialisation.

\begin{figure}[t!]
\centering
\includegraphics[width=0.45\textwidth]{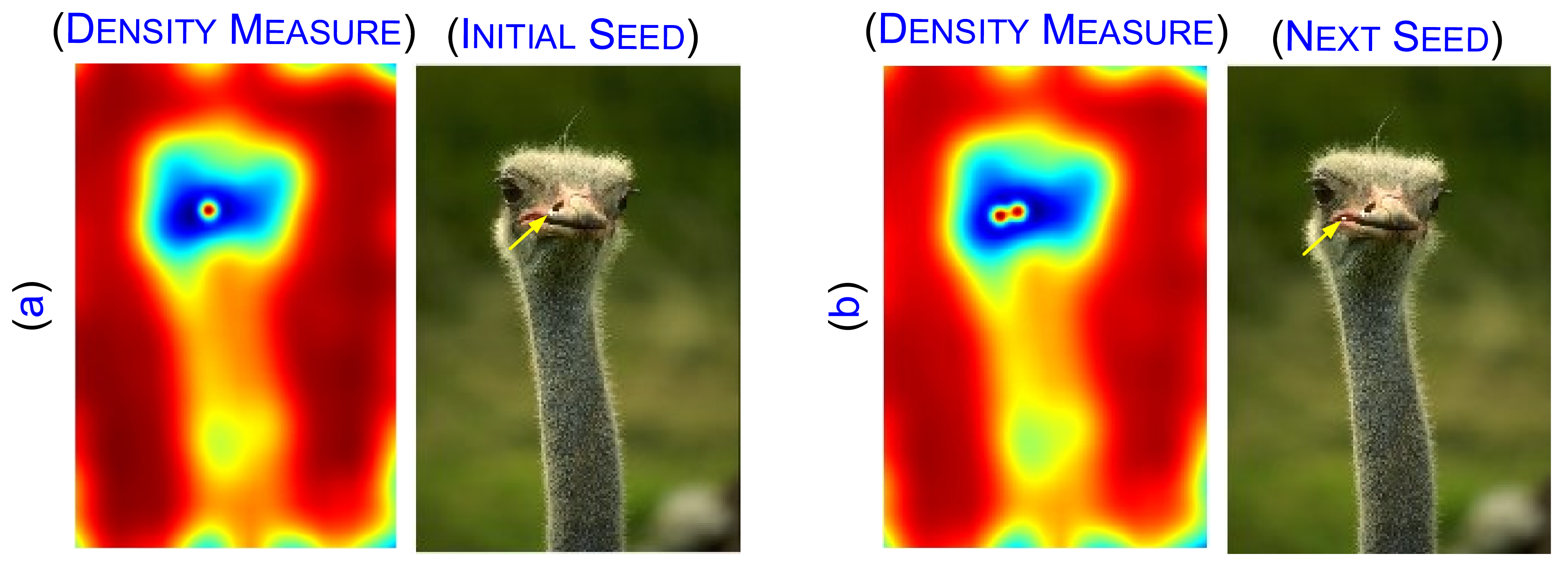}
\caption{Illustration of our initialisation strategy, which incorporates the object density measure $\mathcal{G}$, for two initial seeds.  From left to right, first seed and second one.}
\label{fig::seedingExample}
\end{figure}

\color{black}

\subsection{Seeding Initialisation: A New Strategy}
In this section, we describe our new seeding initialisation. Our main motivation is that we can use the object density measure $\mathcal{G}$ defined above to help seed clusters in object-rich parts of the image, which we expect to contain more distinct regions. In this way, we obtain greater resolution at fewer iterations, and improve the focus on relevant and interesting regions.

We remark to the reader that SLIC initialisation is based on sampling pixels at the image grid. For comparison purposes, we start by defining the SLIC initilisation, which reads as follows.

\begin{shaded*}
\vspace{-0.5cm}
\begin{lstlisting}[language=Python, mathescape=true]
#pseudocode for seeding initialisation SLIC
Set: Initialise cluster centers as
$C_i=[x_i,y_i,l_i,a_i,b_i]$ by sampling at regular grid
step: $S= \sqrt{N/k}$
#where N is the size of the image
Move cluster centers to the lowest gradient
position in a $3\times 3$ neighborhood
\end{lstlisting}
\vspace{-0.5cm}
\end{shaded*}

\noindent
Our proposed approach, which incorporates $\mathcal{G}$ into this seeding, can be described informally as follows. We first set as an initial point the pixel with the lowest  value in $\mathcal{G}$, and then we increase  the values near to the initial point such that its neighbours are unlikely to be selected as another initial point. In this way, we guarantee that the distance between seed points is comparable to the search range, which will help reduce redundant searches. This process is illustrated in Fig.~\ref{fig::seedingExample} for two initial seeds.

\begin{figure}[t!]
\centering
\includegraphics[width=0.5\textwidth]{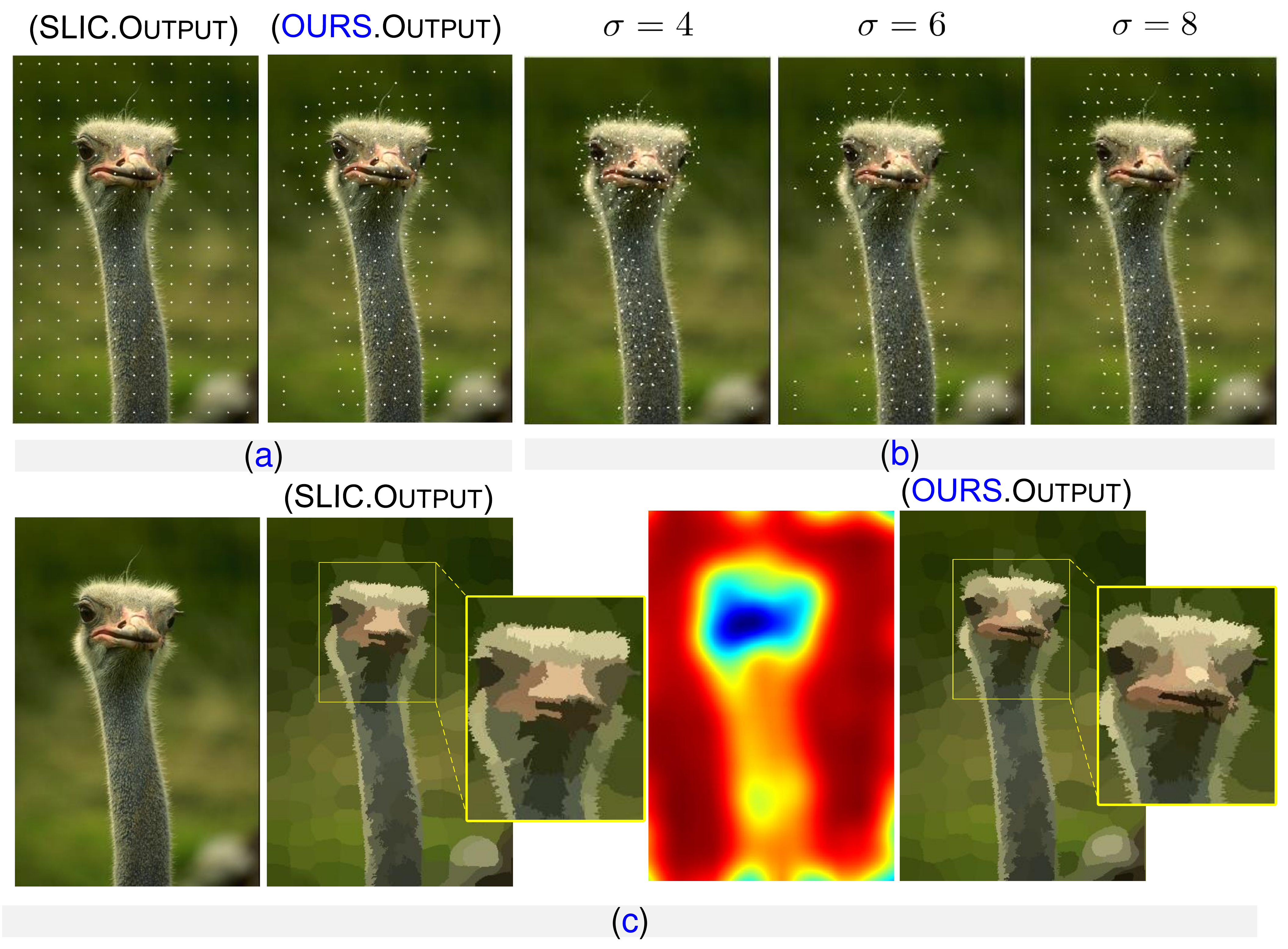}
\caption{Visual comparison of the seeding initialisation of SLIC vs ours.(a) One can see that we seek to focus on relevant areas (i.e. other than background). (b) The effect of $\tau$ in our seeding strategy. {(c) Visual illustration of the positive repercussion of our technique over SLIC. Our approach seeks to focus on structure-rich parts (see the face) whilst avoid over-segmenting uniform areas such as the background. }}
\label{fig::initialisation5}
\end{figure}

The hedging described above is carried out in two stages as follows.
\begin{itemize}
    \item Points adjacent to the initial point are made unselectable, by setting the value of $\mathcal{G}$ at these points excessively large.
    \item Points in the proximity of the initial point are made less likely, but not impossible, to select, again by altering $\mathcal{G}$. The influence range and to what extent are under consideration.
\end{itemize}

\noindent
The advantage of these changes is that the density of area is limited twice compared with original method. The overall procedure of our method, which suitably sets the initialisation points according to our structure measure  $\mathcal{G}$, is described formally as follows.

\begin{shaded*}
\vspace{-0.5cm}
\begin{lstlisting}[language=Python, mathescape=true]
#pseudocode for seeding initialisation OURS
Set $r= max(\mathcal{G})/min(\mathcal{G})$
While Enough Seeds:
    Set range=sqrt(NumOfPixels/NumOfSuperpixels)
    for each Superpixel center $C_j$:
        Initialise $S_i$ with the lowest value
        in $\mathcal{G}$
        Set the adjacent neighbours of $S_i$
        $\mathcal{G}=\infty$
        for all points $S_j$ with $d(S_i,S_j)<range$
            $\mathcal{G}(S_j)=\mathcal{G}(S_j)*sqrt(r)$
            Smooth region with $g_\tau$, $\tau=13/2$

\end{lstlisting}
\vspace{-0.5cm}
\end{shaded*}

\begin{figure*}[t!]
\centering
\includegraphics[width=1\textwidth]{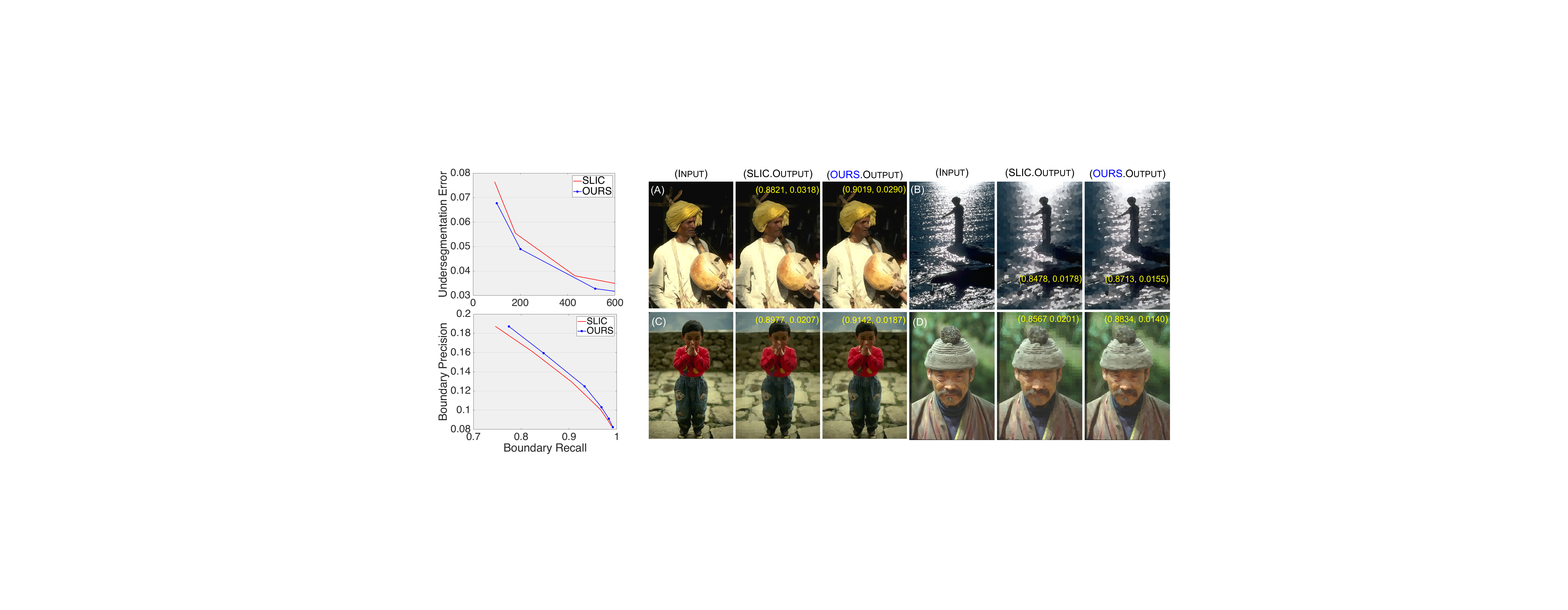}
\caption{From left to right. Quantitative comparison of our approach vs SLIC using three metrics: UE, BR and BP. Our approach reported the best scores metric-wise. Four visual outputs comparisons of SLIC vs our approach. In a closer look, one can see that our approach achieves better connection of structures and keeps fine-details. For example, see (A), (C) and (D) the faces and (B) the hand. {The numerical results in parenthesis denote (Undersegmentation Error, Boundary Recall)}  }
\label{fig::oursvsslic2}
\end{figure*}

An output example is displayed in Fig.~\ref{fig::initialisation5}. Subfigure (a) shows a initialisation comparison between our approach and SLIC, and we see that our approach gives more importance to the ostrich than the background. In subfigure (b), we evaluate possible choices for $\tau$, and display outputs for $\tau= {4,6,8}$. In practise, we found that the $\tau= 13/2$ works for a range of images. {Finally, subfigure (c) shows a visual illustration of the advantage of our technique. By integrating our density function into the distance computation one can avoid unnecessary over-segmentation of large uniform regions (i.e. the background) and aim the attention at structure-rich  parts (i.e. the ostrich). A closer look at the zoom-in regions one can see that our technique was able to preserve fine details in the face whilst SLIC lost relevant areas such as the left eye.}

\section{Experimental Results}
In this section, we describe in detail the experiments that we ran to evaluate our approach.

\subsection{Evaluation Protocol}
\textbf{Dataset Description.} We evaluate our proposed approach  on a publicly available dataset, the Berkeley Segmentation  Dataset~\cite{arbelaez2010contour}, which provides ground truth of the images for quantitative analysis. {Moreover, to further support our technique, we included two interesting cases to study where fine-details are particularly difficult to retrieve: 1) out-of-the-focus images using the dataset of that~\cite{shi2014discriminative} and 2) wide angle images using the dataset from~\cite{yang2019pass}. All  the  measurements  and  comparisons  in  this  section  are  taken  from  these  datasets.}

\smallskip
\textbf{Comparison Methodology.} We compare our approach to the SOTA methods on superpixels. For this, we design a two-part evaluation scheme. For the first part, we compared our approach against SLIC~\cite{achanta2012slic}; this comparison therefore demonstrate that our carefully design solution achieves better performance than the top reference in clustering-based methods. For the second part, we compared to state-of-the-art techniques: QS~\cite{vedaldi2008quick}, TP~\cite{levinshtein2009turbopixels} ,TPS~\cite{tang2012topology}, LRW~\cite{shen2014lazy}, SNIC~\cite{achanta2017superpixels} and dSLIC~\cite{maierhofer2018peekaboo}. We compare our approach qualitatively by visual comparisons and  quantitatively by computing three metrics: Under-segmentation Error (UE), Boundary Recall (BR) and Boundary Precision (BP), which definitions are described next.

\medskip
We assume that we are given an image, along with a \emph{ground truth} $\Gamma=\{g_i\}_{i=1}^M$, representing the true regions of the image.

\textbf{Boundary Recall} measures the proportion of the boundary of the true regions in the ground truth which are close to a boundary in the segmentation. To quantify the notion of being close to a boundary, we recall the following definition.

Given a subset $E$ of the edge set, we define the distance $d(e,E)=\inf\{|e-f|: f\in E\}$, where $|\cdot|$ denotes the $l_2$ norm of the difference, measured in pixels. We then define the \emph{Boundary Recall} as

\begin{mydef}[Boundary Recall] Given a ground truth $\Gamma=\{g_i\}_{i=1}^M$ and a segmentation $\mathcal{S}=\{s_j\}_{j=1}^k$, we write $\partial \Gamma$ for the union of the edge boundaries $\partial g_i$, and similarly write $\partial \mathcal{S}$ for $\{s_j\}_{j=1}^k$. We define the boundary recall by \begin{equation*} \partial(\Gamma, \mathcal{S})=\frac{\#\{e \in \partial \Gamma: d(e, \partial \mathcal{S})\leq 2\}}{\#\partial \Gamma}. \end{equation*} In words, the boundary recall is the proportion of true edges which are close to a superpixel edge.
\end{mydef}

\medskip
\textbf{Undersegmentation Error.} Intuitively, this measures the size of all superpixels which \textit{spill} across boundaries of the ground-truth.
\begin{mydef} For a ground truth $\Gamma=\{g_i\}_{i=1}^M$, we fix thresholds $B_i, i=1, ..,M.$ Given segmentation $\mathcal{S}=\{s_i\}_{i=1}^k$ of the image, the under-segmentation error is given by \begin{equation*}
U=\frac{1}{N}\left[\sum_{i=1}^M \left(\sum_{\#s_j \cap g_i \geq B_i}\#s_j\right)-N\right]
\end{equation*}
Observe that, since $\mathcal{S}$ is a partition of the image, we can rewrite \begin{equation*}
U=\frac{1}{N}\sum_{i=1}^M\left[\left(\sum_{\#s_j \cap g_i \geq B_i}\#s_j\right)-\#g_i\right]
\end{equation*} Hence, the undersegmentation error $U$ measures how wasteful the coverings of the true regions $g_i$ by the superpixel regions $s_j$ are. We usually take $B_i$ to be a fixed proportion of $\#g_i$
\end{mydef}

\smallskip
\textbf{Parameter Selection.} For all compared approaches QS~\cite{vedaldi2008quick}, TP~\cite{levinshtein2009turbopixels}, TPS~\cite{tang2012topology}, SLIC~\cite{achanta2012slic}, LRW~\cite{shen2014lazy}, SNIC~\cite{achanta2017superpixels} and dSLIC~\cite{maierhofer2018peekaboo}, we set the parameters as suggested in the corresponding work. We also used the codes released from each corresponding author. For our approach,  we set the $m=10$ since it offers a good trade-off between shape uniformity and boundary adherence (see Supplementary Material Section 2 for further description on $m$). We performed the evaluation using up to a range of number of superpixels up to 600.

\Angie{The experiments reported in this section were under the same conditions in a Matlab CPU-based implementation. We used an Intel Core i7 with 16GB RAM. }

\begin{figure*}[t!]
\centering
\includegraphics[width=1\textwidth]{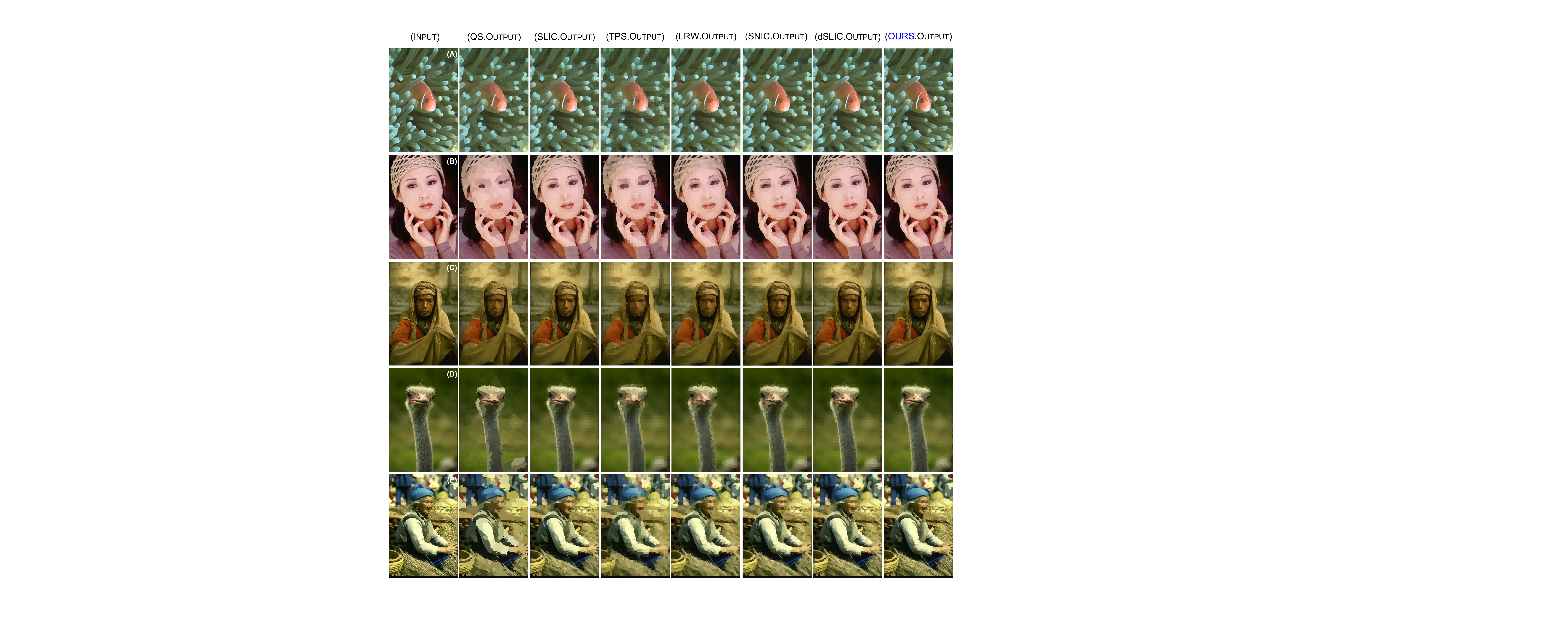}
\caption{Superpixel outputs comparisons of our approach vs different methods from the body of literature: QS~\cite{vedaldi2008quick}, SLIC~\cite{achanta2012slic} TP~\cite{levinshtein2009turbopixels}, TPS~\cite{tang2012topology}, LRW~\cite{shen2014lazy}, SNIC~\cite{achanta2017superpixels} and dSLIC~\cite{maierhofer2018peekaboo}. A closer inspection, one can see that our approach offers better superpixels segmentation. For example, (A), (B), and (C) the eyes; (D) the ostrich's boundary and (E) the eyes and basket. }
\label{fig::compALL1}
\end{figure*}

\begin{figure*}[t!]
\centering
\includegraphics[width=1\textwidth]{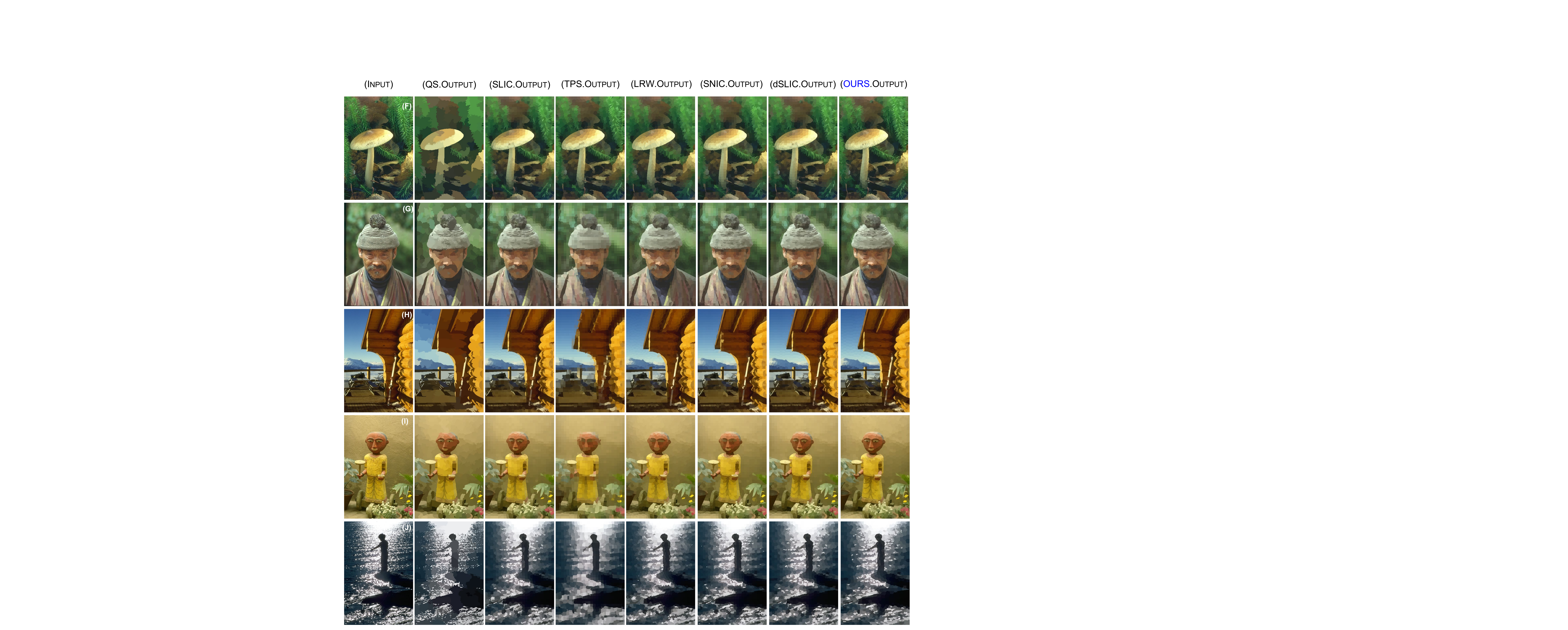}
\caption{Superpixel segmentation outputs of our approach vs QS~\cite{vedaldi2008quick}, SLIC~\cite{achanta2012slic} TP~\cite{levinshtein2009turbopixels}, TPS~\cite{tang2012topology}, LRW~\cite{shen2014lazy}, SNIC~\cite{achanta2017superpixels} and dSLIC~\cite{maierhofer2018peekaboo}.
Visual assessment shows that the proposed algorithm performs better than the compared approaches. Examples are: (F) the leaves; (G) and (I) the face; (H) the house boundaries and (J) the hand.}
\label{fig::compALL2}
\end{figure*}

\subsection{Results and Discussion}
We divide this section in two parts, following the comparison methodology scheme presented in previous sections.  \smallskip

\textbf{$\triangleright$ Is our Approach better than SLIC?}  As SLIC approach remains a top reference, and is the basis of our approach, we start by evaluating our approach against it. Results are displayed in Fig.~\ref{fig::oursvsslic2}. In a closer look, at the right side, of this figure, one can see that our approach yields to a better segmentation of the structures, keeping fine details of the objects. Moreover, it avoids unnecessary oversegmentation on uniform areas. These positive properties of our approach can be seen, for example, in  (B) the proper recovery of the hand; in (C) the hair, eyebrows and the lines patterns in the jumper that are correctly clustered; in (D) where our approach successfully capture the eyes and moustache, and in (A) with better preservation of the face structure including the nose and lips.

To further support of our visual results, we ran a quantitative analysis based on three metrics UE, BR and BP.  The results are displayed at the left side of Fig.~\ref{fig::oursvsslic2}. \Angie{The top part shows a comparison in terms of UE, where the results reflect conformity to the true boundaries. We can observe that our approach achieves the lowest UE for all superpixels counts. The same positive effect was found in terms of precision versus recall, in which our approach displayed the best performance. Overall, our proposed technique outperforms SLIC at all superpixels counts and reduces the error rate by 20\%. This improvement is translated to our approach to be the best in terms of producing superpixels that respect the object boundaries.}

\begin{figure*}[t!]
\centering
\includegraphics[width=1\textwidth]{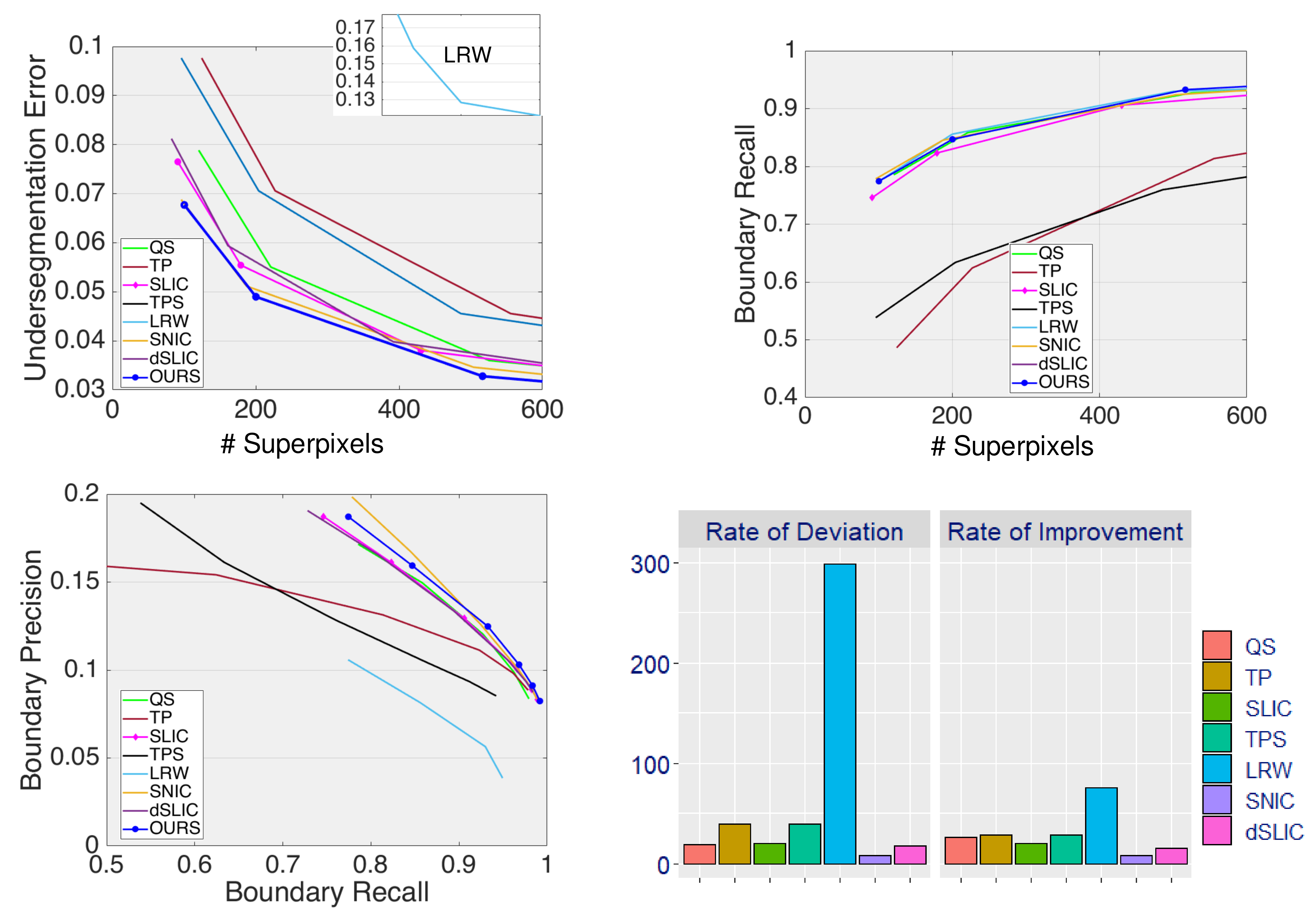}\vspace{-0.5cm}
\caption{Metric-wise comparison of our approach vs SOTA techniques using UE, BR and BP. In a closer look, we can see that our approach, overall, offers the lowest UE and the highest BR. Finally, the good boundary adherence to the true edges is reflected in the last plot, in which our approach overall gets the best trade-off between those metrics. \Angie{These results are further supported by the rate of deviation and rate of improvement.} }
\label{fig::plotsALL}
\end{figure*}

\begin{figure*}[t!]
\centering
\includegraphics[width=1\textwidth]{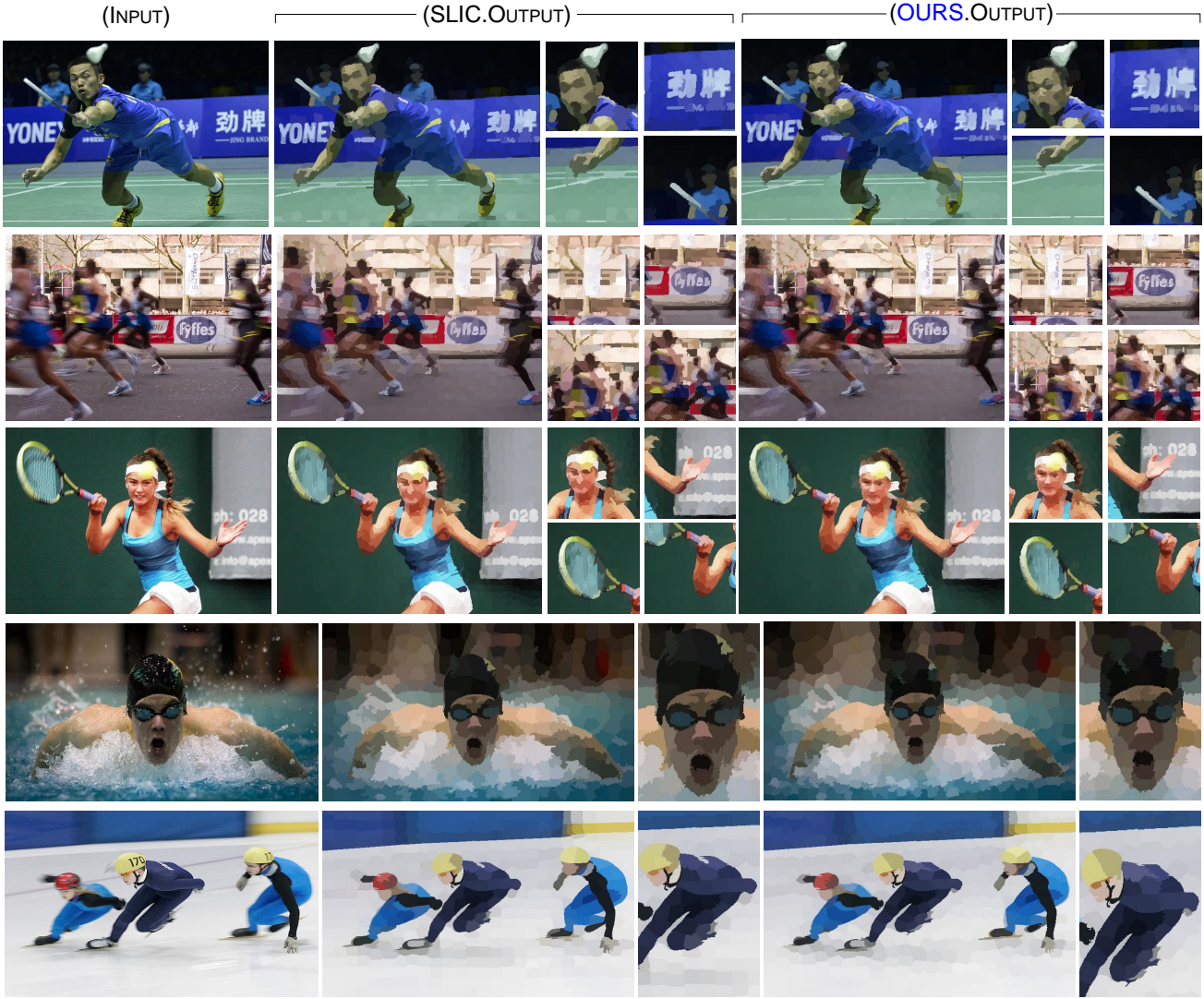}
\caption{\Angie{Visual comparison of our approach and SLIC using out-of-the-focus sport images. Zoom-in views display relevant areas, in which our approach successfully retrieve fine detail and blurry structures.}}
\label{fig::fig10}
\end{figure*}

\begin{figure*}[t!]
\centering
\includegraphics[width=1\textwidth]{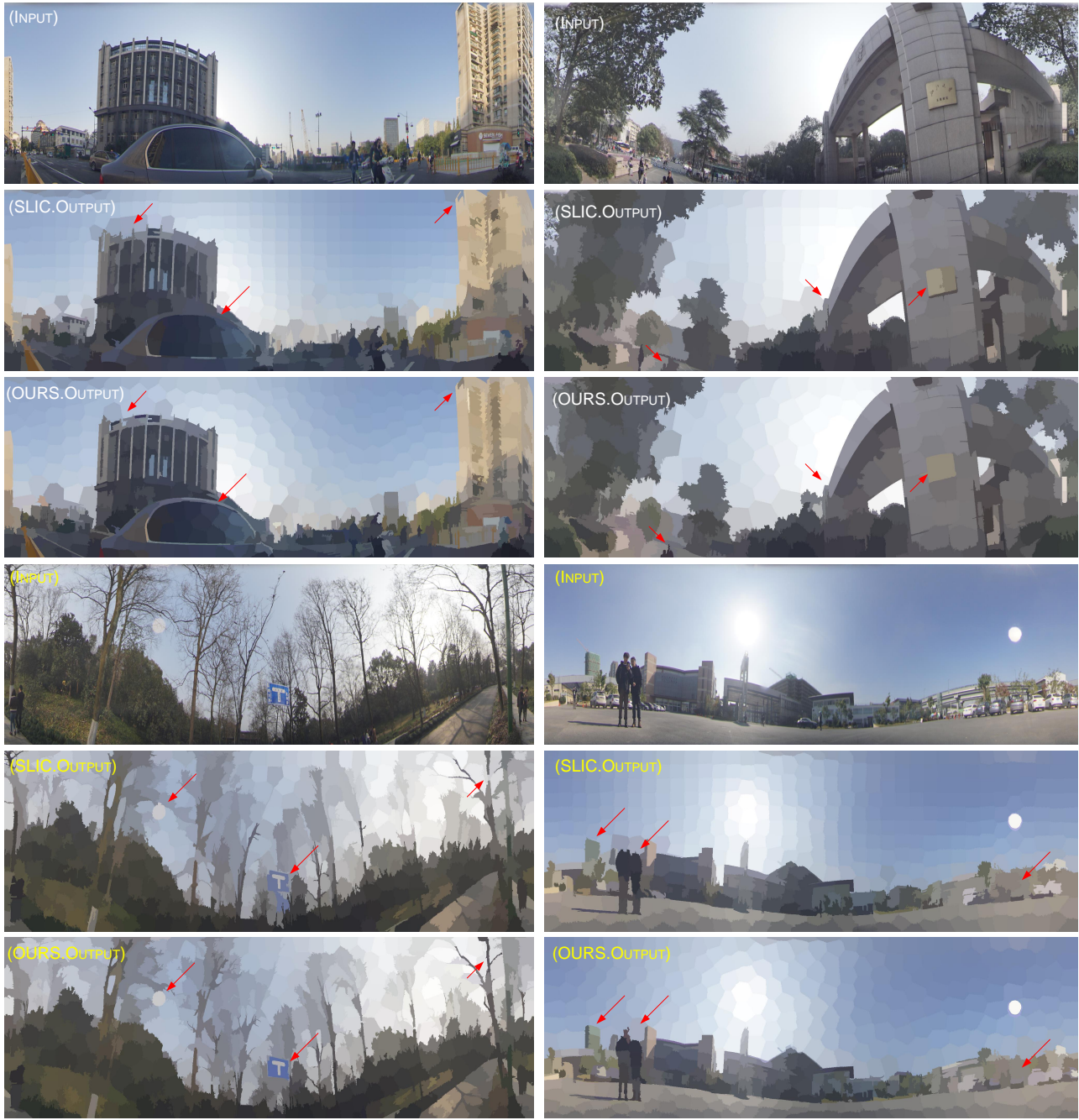}
\caption{\Angie{Visual assessment of our techique vs SLIC using wide angle images, in which the objects are distorted and  the segmentation task becomes more challenging. Red arrows show interesting areas, in which our technique improves over SLIC.}}
\label{fig::fig12}
\end{figure*}

\begin{figure}[t!]
  \begin{center}
   \includegraphics[width=0.45\textwidth]{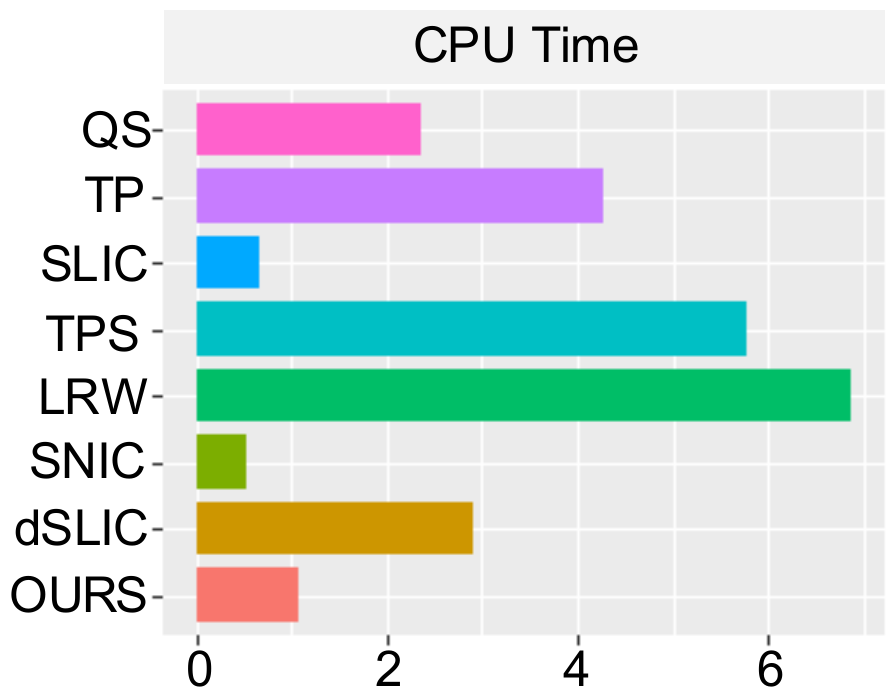}
  \end{center}
  \caption{CPU time averaged comparison of our approach vs the body of literature. One can see that our approach improvement comes at a negligible cost in runtime in comparison with the fastest approaches SLIC and SNIC.}
  \label{fig::CPUtime}
\end{figure}

\smallskip
\textbf{$\triangleright$ Is our approach better than other Superpixel approaches? } As the second part of our evaluation, we compare our approach against SOTA models:  QS~\cite{vedaldi2008quick}, TP~\cite{levinshtein2009turbopixels}, TPS~\cite{tang2012topology}, LRW~\cite{shen2014lazy}, SNIC~\cite{achanta2017superpixels} and dSLIC~\cite{maierhofer2018peekaboo}. We selected for our comparison approaches coming from different perspectives: graph-based, path-based, density-based and clustering-based approaches. Results are displayed in Figs. ~\ref{fig::compALL1}, \ref{fig::compALL2} and \ref{fig::plotsALL}.

We first present a visual comparison of a selection of images from the Berkeley dataset in Figs. \ref{fig::compALL1} and \ref{fig::compALL2}. By visual inspection, one can see that QS and TPS are the ones that perform more poorly than the other compared approaches. They fail to provide good boundaries of the structures and they do not preserve relevant details. Examples can be seen in (A), (B), (E) and  (G) the eyes; (C), (E) and (F) preservation of fine details. LRW offer a better edge adherence to the objects than QS and TPS but also fails to preserve relevant objects, for example (G) the eyes and moustache.

In contrast to those approaches, SLIC and dSLIC performs better than QS, TPS and LRW. One can observe that SLIC and dSLIC readily compete in terms of having better boundary adherence to the structures and grouping correctly majority of the objects. However, in particular, SLIC still produces outputs with more superpixels than necessary in homogeneous parts of the structures; see, for instance, in (A) the fish eye, (B) the nose and (J) the hand. Although dSLIC performs slightly better than SLIC, it still fails to capture fine details.

Among those approaches, SNIC approach display more robustness in terms of grouping structures correctly than the previous approaches. However,  like SLIC it also tends to generate more superpixels than needed in uniform regions so that the final  outputs do not capture fine details. Examples are (G), (E) and (C) face details; (I) the eyes and head  and (F) the leaves.

These major drawbacks are mitigated by our model. Our algorithmic approach shows the best boundary adherence and regularity. This is visible in the leaves in image (F), in which our approach is able to better capture the structure, in (I) on the lips where our approach is able to capture the correct geometry, and (A) the fish eyes, where our approach is the only one that correctly segments the inner part. These positive properties of our approach are prevalent in all images. More examples include preservation of the geometry such as in (J) the hand and (G) face, in which our model is the only one able to correctly segment these fine details.

To further support our visual evaluation, we show a metric-wise comparison in Fig.~\ref{fig::plotsALL}. We start by evaluating the approaches in terms of UE, which is displayed at the left side of this figure. Close observation shows that QS, TPS and TP perform poorly, and in particular, LRW that reported the highest Undersegmentation Error. dSLIC and SLIC show quite low UE, and SNIC ranks the second best. Overall, our approach shows substantial improvement over the compared approaches reporting the lowest UE for all superpixels counts.  A similar effect is exhibited in terms of Boundary Recall. TP and TPS perform poorly while the other compared approaches reported better BR. Our approach readily competes with the other compared schemes and the overall BR of our approach was reported to be the best. The same effect is observed in terms of BP-vs-BR, which reflects that our approach overall adheres better to the truth boundaries. \Angie{To have a better sense of the improvement, we finally report the rate of deviation and rate of improvement of all compared techniques with respect to ours, which results are displayed at the bottom left side of Fig.~\ref{fig::plotsALL}. With respect to the rate of deviation our technique displays a clear difference ranging from ~299\% to 8\% whilst in terms of rate of improvement a range of ~75\% to 10\%. In particular, the worst performance was reported by LRW in which we improved substantially with  299\% and 75\%  in the rate of deviation and improvement. Our closest competitor is SNIC with which we reduce the error rate by more than 10\%. This improvement comes at a negligible cost time which can be seen in Figure~\ref{fig::CPUtime}.}

\textbf{$\triangleright$ How is the Computational Performance?} Finally, we evaluate our approach vs the SOTA models in terms of CPU performance in seconds. Results are displayed at Figure ~\ref{fig::CPUtime}, using the average time across all images and over the range of [80, 2500] superpixels. From this plot, we observe that TP, TPS and LRW require high computational time whilst QS and dSLIC sightly  improve in this regard. Finally, SLIC, SNIC and OURS provide more feasible runtimes that are appropriate for a pre-processing task. \Angie{We remark that our negligible  computational load   ($\sim$5\%) is  justified by the substantially improved results over SLIC, SNIC with the same number of superpixels. Overall, the computational demand of our technique is of practical interest for several domains, where  decreasing the computational load by reducing the number of primitives in the image domain is needed, including remote sensing, medical applications and video analysis. }

\begin{figure}[t!]
\centering
\includegraphics[width=0.5\textwidth]{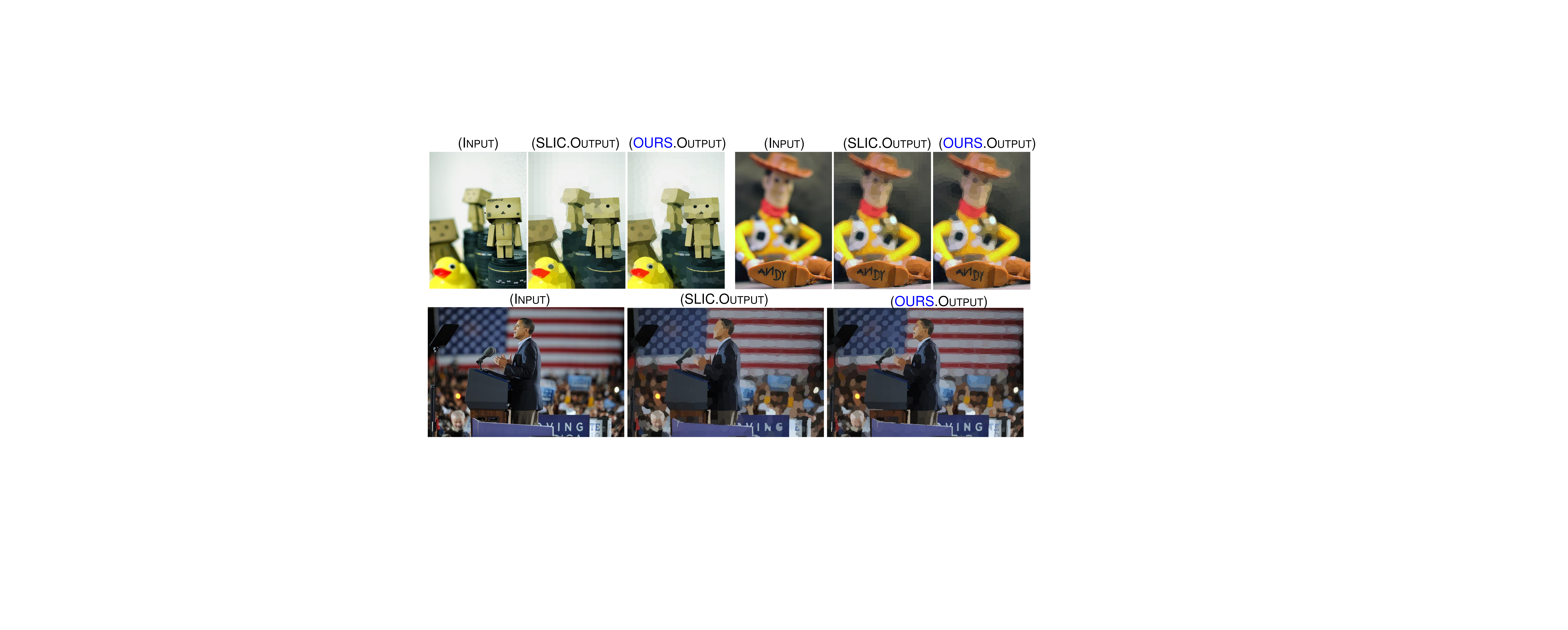}
\caption{\Angie{Visual comparison against SLIC using out-of-the-focus images. The selected frames show challenges cases when objects are highly blurry. Overall, our technique gives a better superpixel segmentation by preserving objects structures and fine details. }}
\label{fig::fig11}
\end{figure}

\textbf{$\triangleright$ Case Study:  Out-of-the Focuss  and Wide Field of View Images }
\Angie{To further support our technique, we included two interesting yet challenging cases. The first one is superpixels with out-of-the focus images. Whilst the second case is when using wide field of view images.

We start evaluating our approach using out-of-focus images and the visual outputs of our technique and SLIC are displayed in Figures  \ref{fig::fig10} and \ref{fig::fig11}. The study of this case is relevant to assess the effectiveness of our technique in retrieving fine details and rich-structures with  blurry and non blurry objects in the image. A common scenario of this case is in sport events due to the fast actions in the scene. In Figure \ref{fig::fig10}, we displayed a set of outputs and in a closer look, one can see that our technique was able to preserve relevant structures. We provide zoom-in snapshots of interesting cases. For example see the face details in all image, in which our technique segmented correctly the eyes, mouth and nose. Another example is the letters in the background from the first, second and third images. We also included a set of different images with different scenarios in Figure \ref{fig::fig11}, in which previous positive effects are also observed. For example at the first image the left eye is correctly retrieved with our technique whilst SLIC failed to segmented correctly. We observed that blurry parts are also well-segmented by our technique, and in particular, also in keeping fine details. Overall, the outputs displayed in Figures \ref{fig::fig10} and \ref{fig::fig11} suggest that our proposed method produces visually more pleasant results than SLIC.

Our second case of study is when using wide angle images. The challenging when segmenting is due to the structure of the objects become highly distorted  with respect to the distance between the camera point and the object position in the scene. This makes image segmentation harder than rectilinear images. We then investigate if our technique is still efficient in terms of keeping the object shape and fine details. With this purpose, we ran our technique and SLIC over a set of image samples from ~\cite{yang2019pass}, and the results are reported in Fig. \ref{fig::fig12}. A closer look at the outputs, one can observe that our technique fulfill its purpose of  avoiding over-segmenting homogeneous parts such as the sky and focus on the rich parts for example buildings or people. Visual comparison of this effect can be seen as pointed out by the red arrows. Overall, our technique shows potentials for the two investigated cases by generating better output segmentations in terms of retrieving fine details and preserving the objects shape.}

\section{Conclusion}
In this work, we proposed a new superpixel approach that builds upon SLIC technique. Our approach incorporates the notion of spectral residual as a proxy for object density and a novel seeding strategy. We demonstrated that our approach seeds clusters advantageously and modifies the local search radius. This leads to better segmentation, with a comparable computational load, to other state-of-the-art algorithms.
These modifications leads to major advantages in terms of avoiding the over-segmentation of uniform areas in terms of fine-detail preservation, with a comparable computational load to other state-of-the-art algorithms.

\Angie{We offer the following two remarks on how future work may reduce the computational load of our technique. First, we observe that our slightly higher computational load is justified by the substantially improved results over SLIC and SNIC with the same number of superpixels. This suggests that, by reducing the number of superpixels in our approach, we could achieve better segmentation than SLIC and SNIC while also reducing computational load. Secondly, we remark that the full saliency measure $\mathcal{SR}(x)$ contains more information that is strictly necessary for our technique, since it is only used as a \emph{rough} measure of object density, while also adding substantially to the computational load. We therefore suggest that a density measure $\mathcal{G}$ could instead be constructed from a \emph{down-sampled} version of the image so that the computational load could be significantly reduced, while maintaining excellent segmentation. For instance, down-sampling by a factor of $2$ will reduce the computational load of computing $\mathcal{SR}$ by a factor of $4$ and, provided that the superpixel resolution is much lower than the the resolution of the true image, have minimal effect on the initial seeding. We also note that the map $\mathcal{SR}$ used here is certainly not the only possible technique for determining object density; future research may consider using other techniques, possibly based on machine learning, to replace our object density measure $\mathcal{G}$. Furthermore, future work will include  performance evaluation over a range of applications, where superpixels makes significant positive effect in decreasing the computational cost. For example, in remote sensing, video analysis and medical image analysis. }

%
%


\section*{Acknowledgments}
AIAR gratefully  acknowledges  support  from  CMIH (EP/N014588/1 and EP/T017961/1)  and CCIMI, University of Cambridge. DH is supported by the UK Engineering and Physical Sciences Research Council (EPSRC) grant EP/L016516/1 for the University of Cambridge Centre for Doctoral Training, the Cambridge Centre for Analysis.
XZ was partially supported by the Research Grants
Council of Hong Kong (Project No. CityU 11301419) and City University
of Hong Kong (Project No. 7005497).
RC's research is supported by HKRGC Grants No. CUHK 14306316 and CUHK 14301718, CityU Grant 9380101, CRF Grant C1007-15G, AoE/M-05/12.
CBS acknowledges support from the Leverhulme Trust project on 'Breaking the non-convexity barrier', the Philip Leverhulme Prize, the Royal Society Wolfson Fellowship, the EPSRC grants EP/S026045/1 and EP/T003553/1, EP/N014588/1, EP/T017961/1, the Wellcome Innovator Award RG98755, European Union Horizon 2020 research and innovation programmes under the Marie Skodowska-Curie grant agreement No. 777826 NoMADS and No. 691070 CHiPS, the CCIMI and the Alan Turing Institute.

{\small
\bibliographystyle{ieee}
\bibliography{mainSub}
}

\end{document}